\documentclass[sigconf]{acmart}

\usepackage{amsmath}
\DeclareMathOperator*{\argmax}{argmax}
\usepackage{graphicx}
\usepackage{dcolumn}
\usepackage{arydshln}
\usepackage{verbatimbox}
\usepackage{balance}
\usepackage{mathrsfs}  

\usepackage{setspace, graphicx, fullpage, fancyhdr, amsmath, epsfig, natbib, array, multirow, hyperref, tabularx, lscape, booktabs, sidecap, subfig, longtable, enumitem}
\usepackage{amsmath,amsfonts,amsthm, bm}
\usepackage{caption} 
\AtBeginDocument{%
  \providecommand\BibTeX{{%
    \normalfont B\kern-0.5em{\scshape i\kern-0.25em b}\kern-0.8em\TeX}}}

\begin{document}

\title{Accounting for Variations in Speech Emotion Recognition with NonParametric Hierarchical Neural Network}


\vspace{1cm}

\author{Lance Ying}
\affiliation{%
  \institution{University of Michigan}
  \city{Ann Arbor}
  \state{Michigan}
  \country{USA}
}

\author{Amrit Romana}
\affiliation{%
  \institution{University of Michigan}
  \city{Ann Arbor}
  \state{Michigan}
  \country{USA}}

\author{Emily Mower Provost}
\affiliation{%
  \institution{University of Michigan}
  \city{Ann Arbor}
  \state{Michigan}
  \country{USA}}


\begin{abstract}
In recent years, deep-learning-based speech emotion recognition models have outperformed classical machine learning models. Previously, some neural network designs, such as Multitask Learning, have accounted for variations in emotional expressions due to demographic and contextual factors. However, existing models face a few constraints: 1) they rely on a clear definition of domains (e.g. gender, culture, noise condition, etc.) and the availability of domain labels. 2) they often attempt to learn domain-invariant features while emotion expressions can be domain-specific. In the present study, we propose the Nonparametric Hierarchical Neural Network (NHNN), a lightweight hierarchical neural network model based on Bayesian nonparametric clustering. In comparison to Multitask Learning approaches, the proposed model does not require domain/task labels. In our experiments, the NHNN models outperform the models with similar levels of complexity and state-of-the-art models in within-corpus and cross-corpus tests. Through clustering analysis, we show that the NHNN models are able to learn group-specific features and bridge the performance gap between groups.

\end{abstract}




\maketitle
\section{Introduction}

Emotion can greatly influence people’s mental processes and behaviors and is also a predictor of physical and psychological well being. Speech Emotion Recognition (SER) has become an important research topic in human-computer interaction. 

Although considerable research effort has been put into developing models capable of recognizing and predicting human emotions, much of the work focuses on lab-produced datasets\cite{busso2008iemocap} and robust speech emotion recognition on in-the-wild speech with diverse confounding acoustic elements remains a challenge \cite{zhang2017cross}. Previous work has used a variety of machine learning techniques such as multi-task learning \cite{zhang2017cross} and domain generalization \cite{yin2020speaker} that attempt to learn a more robust and generalizable representation of speech features, yet few studies have successfully accounted for variations in speech emotion corpora due to the complexity of emotion expressions and a variety of moderating variables such as gender, culture, etc. 

In this paper, we propose a novel lightweight architecture based on unsupervised non-parametric clustering and supervised convolutional neural networks (CNN) in order to account for inter-group differences in emotional expressions. We test the model on both lab-produced and in-the-wild speech emotion datasets and show our proposed model outperforms state-of-the-art methods on most within-corpus and cross-corpus tests. Our work has the potential to achieve robust SER performances in groups with a variety of emotional expressions.

\section{Related Work}

\subsection{Speech Emotion Recognition}

Speech Emotion Recognition (SER) has attracted enormous interest in the past decades.  Past research has experimented with different machine learning algorithms, such as Hidden Markov Models (HMM)\cite{mao2019revisiting}, Artificial Neural Networks (ANN) \cite{chen2012speech}, Support Vector Machines (SVM) \cite{jin2015speech}. In recent years, deep learning approaches have become more popular and have reached state-of-the-art performances on SER  \cite{gideon2019improving}. For instance, Zhao et al. proposed the CNN-LSTM model, which showed better accuracy than conventional CNN and LSTM models~\cite{zhao2019speech}. Liu et al. proposed a temporal attention CNN model, which shows stable and state-of-the-art performances on SER tasks~\cite{liu2020temporal} . 

\subsection{Acoustic Features and Variations}

A considerable amount of research from the speech signal processing community has explored relevant acoustic features for SER \cite{sudhakar2015analysis}. Some common features used for classifications often include prosody features, such as pitch and  loudness, voice quality features, such as the first three formants and the signal-to-noise ratio, and spectral features, such as the Melfrequency Cepstral Coefficients (MFCC) and Linear Prediction Cepstral Coefficients (LPCC).

However, robustly extracting emotion from these acoustic features can be quite challenging. The relationship between experienced emotions and observed acoustic features can be moderated by multiple contextual and demographic factors. These variations create challenges for robust automatic speech recognition systems. One source of variations is environmental conditions. In-the-wild speech is often recorded with different recording devices, producing different sound qualities. A variety of background noise, such as other human speech and music, can also introduce noise into the extracted acoustic features. Additionally, researchers have shown that the experience of emotion and its expressions are often influenced by biological, social, and cultural factors. Kring and Gordon found that, when presented with emotional stimuli, female participants were more expressive than male participants, even though they reported the same type and intensity of experienced emotions \cite{kring1998sex}. Matsumoto et al. found that cultural display rules differ systematically between Japan and the United States \cite{matsumoto1990cultural}. Gender differences have also been well documented in SER applications. For instance, Alghowinem et al. found that best features for depression detection from speech differed across genders \cite{sharifa2012joyous}.

Despite advances in speech emotion recognition systems, few model designs have properly accounted for these factors. One popular approach is to partition audio segments to be part of a certain domain, such as gender, corpus, noise condition, etc., and apply domain adaptation \cite{wilf2020dynamic} or multi-task learning during training \cite{zhang2017cross}. Some have trained separate models on each domain and then applied model fusion in a later stage \cite{schuller2011using}. However, one key assumption of such approaches is that there exist pre-specified domains for audio. For human speech, the domains can be non-exhaustive. For instance, gender, race, language, age, environmental conditions are all possible criteria for categorization, yet a combination of these criteria would generate an exponential amount of permutations with multiple labels (e.g. Male, Caucasian, adult, etc.). Training separate models for each permutation or applying domain adaptation among such a large number of domains are simply impractical.

In addition, the goal of domain adaptation approaches is to find some shared representation that exists across domains. However, some acoustic features may be domain-specific. Past research has shown that negative transfer can occur where transfer methods decrease performances \cite{torrey2010transfer}.

Last but not least, working with pre-specified domains assumes sufficient information about the corpus. However, when training on in-the-wild speech or real-world speech recordings, researchers often don't have explicit and perfect knowledge on subjects' demographics or speech recording conditions, and many of these labels may be incomplete. 

In the present study, we introduce the Bayesian Nonparametric method (BNP) as a way to account for acoustic variations. Our proposed method offers a more flexible alternative than Multi-task Learning or domain adaptation as it captures the latent structure of the dataset through unsupervised means and does not rely on any additional labels except ground-truth emotion categories. The number of domains can grow as new data becomes available.

\subsection{Bayesian Nonparametric Method}

\subsubsection{Dirichlet Process and Dirichlet Process Mixture Model}

The Bayesian nonparametric (BNP) method was first proposed by Ferguson \cite{ferguson1973bayesian}, who introduced the Dirichlet process, denoted as DP($\alpha_0, G_0$), where $\alpha_0$ is the scaling parameter, and $G_0$ is the probability measure.  An explicit formulation was provided by Sethuraman \cite{sethuraman1994constructive} with the famous “stick-breaking construction”, which uses an independent sequence of i.i.d. random variables $\boldsymbol{\beta}=(\beta_1,\beta_2,...)$ and components $\boldsymbol{\phi}=(\phi_1,\phi_2,...)$ where
\begin{equation}
\begin{split}
    &\beta_k | \alpha_0, G_0 \sim Beta (1, \alpha_0) \\
    &\phi_k | \alpha_0, G_0 \sim G_0 
\end{split}
\end{equation}

Then we can define a discrete distribution $G$ as

\begin{equation}
\begin{split}
    &\pi_1=\beta_1, \pi_k=\prod_{i=1}^{k-1}{(1-\beta_i)\beta_k} \\
    &G=\sum_{k=1}^{\infty}{\pi_k\delta_{\phi_k}}
\end{split}
\end{equation}
where $\delta_{\phi_k}$ is a probability measure concentrated at $\phi_k$.

One of the most popular application of the Dirichlet process is to serve as a nonparametric prior for the latent mixing measure in a mixture model, which can be expressed as follows:
\begin{equation}
\begin{split}
    &\theta_i|G \sim G \\
    &x_{i} | \theta_i \sim F(\theta_i) 
\end{split}
\end{equation}
where $x_i$ is the observation, and $F(\theta_i)$ denotes the distribution of $x_i$ given $\theta_i$. This is referred as the Dirichlet Process Mixture Model (DPMM). Since $G$ can be represented with the stick-breaking construction process defined earlier, we can express $\theta_i$ with atoms $\phi_k$. This results in an alternative representation:
\begin{equation}
\begin{split}
\begin{aligned}
    &\boldsymbol{\pi}|\alpha_0 \sim GEM(\alpha_0) &&& \phi_{k} | G_0 \sim G_0 \\
    &z_i|\boldsymbol{\pi} \sim \boldsymbol{\pi} &&& x_i|z_i,\boldsymbol{\phi}\sim F(\phi_{z_i})
\end{aligned}
\end{split}
\end{equation}
Based on this formulation, Antoniak \cite{antoniak1974mixtures} shows that the likelihood of the number of mixtures $k$, given $n$ observations, is calculated to be
\begin{align}
    p(k|\alpha_0,n)=\frac{n!\Gamma(\alpha_0)}{\Gamma(\alpha_0+n)}z_{nk}\alpha_0^{k} \propto z_{nk}\alpha_0^{k}
\end{align}
where $z_{nk}$ is the unsigned Stirling number of the first kind. The expected number of mixtures is 
\begin{align}
    E[k|\alpha_0,n]=\sum_{k=1}^{n}{kp(k|\alpha_0,n)}=\alpha_0\sum_{k-1}^{n}{\frac{1}{\alpha_0+k-1}} \approx \alpha_0\ln{\left(\frac{n+\alpha_0}{\alpha_0}\right)}
\end{align}

In comparison to a standard Gaussian Mixture Model (GMM), the DPMM is able to infer the number of components from the data where in a GMM, the number of components is a predetermined hyperparameter. DPMM is preferred in our study because the groups with variations in acoustic emotion expression are latent. The DPMM can account for such latent structures within the acoustic data.

\subsubsection{Inference of DPMM}

Given observations, the inference of the Dirichlet Process Mixture Model often uses one of the two methods: Markov Chain Monte Carlo \cite{neal2000markov} and Variational Inference \cite{blei2006variational}. In our experiments, the inference of the mixtures is estimated by BNP packages in sklearn \cite{scikit-learn}, which uses the Variational Inference method.

\subsubsection{Applications of BNP}

Due to the variable and complex nature of human behaviors, BNP has been commonly applied to model human activities. Navarro et al. introduced a framework with the Dirichlet Process to model individual differences in category learning, publication habits of psychologists, and web browsing activities \cite{navarro2006modeling}. Chen et al. applied the Dirichlet Process Mixture Model to learn Pedestrian Motion Patterns \cite{chen2016predictive}. These past research showed that BNP was a powerful tool that could be capable of capturing the similarities and differences among emotional domains.

Some previous work has incorporated BNP in SER. For instance, Wang et al. used a hierarchical Dirichlet Process mixture model on music emotion labeling and showed better performance on music emotion annotation and retrieval than GMM \cite{wang2015hierarchical}. However, the overall application of BNP in SER is quite limited and none has attempted using BNP in modeling individual/group differences in corpora.

\section{Classification Model}

In this section, we present the baseline models and the proposed Nonparametric Hierarchical Neural Network.

\subsection{Baseline Models}
\subsubsection{Dilated Convolutional Neural Network (DCNN)}

Convolutional Neural Network(CNN) is one of the most popular deep learning methods. In many applications, CNN has well surpassed the performances of classical machine learning methods because of its ability to construct rich representations from data \cite{khorram2017capturing}. 

The CNN model consists of a feature encoder, which is a stack of convolutional and maxpool layers, and a classifier, which is a stack of fully-connected layers and a softmax layer. For classifications, the classifier generates an n-class probability distribution. 

In the present study, we use a Dilated Convolutional Neural Network (DCNN), which uses two 1D dilated convolutional layers as the feature encoder for inputs. In comparison to a CNN, the dilation in the DCNN allows for a more inclusive receptive field, which is commonly used in SER \cite{khorram2017capturing, kwon2021mlt}. We use ReLU as the activation function, except the softmax at the end. The channel size for the convolutional layers and fully-connected layers are both 128, as used in \cite{khorram2017capturing}.


\subsubsection{Multitask CNN (MTL-CNN)}

Multitask Learning attempts to learn models that perform well on multiple tasks simultaneously. Through this paradigm, models can learn shared representations for multiple tasks with better generalizability. In a simple Multitask CNN, different tasks share the same feature encoder while having task-unique classification layers. We include Multitask CNN as a baseline because it also attempts to account for variations in speech domains and its architecture have a similar amount of parameters and training time as our proposed methods. In the present study, we follow existing work \cite{zhang2017cross,nediyanchath2020multi} and use gender classification as an auxiliary task in addition the main task of valence classification. 

\subsubsection{CNN with Long Short-Term Memory (CNN-LSTM)}

Zhao et al. \cite{zhao2019speech} proposed a CNN-LSTM model which achieved state-of-the-art performance on SER tasks. In this paper, we use the same 1D CNN-LSTM model, which consists of 4 convolutional layers, each followed by a max pooling and dropout layer, 1 LSTM layer and a fully-connected layer.

\subsection{Nonparametric Hierarchical Neural Network}

The NHNN model (Fig. 1) is a hierarchical model that consists of a shared feature encoder and domain-specific classifiers. The feature encoder aims to learn a shared representation for all domains while the domain-specific classifier learns a domain-specific mapping from the bottleneck features to outputs. 

Once the model is trained, the model performs the classification task as follows.Suppose the classification problem has $n$ classes and assume $k$ mixture components are inferred from a Gaussian DPMM with parameters $\boldsymbol{\mu}$, $\boldsymbol{\sigma}$ and cluster weights $\boldsymbol{\omega}$. Denote the set of mixtures as $\boldsymbol{\phi}=\left(\phi_1,\phi_2,...,\phi_k\right)$, the observation as $x$, the class of the observation as $z$, and the latent discrete allocation variable as $\theta$. Each of the classifiers produce a probability distribution $\pi(\phi_i)=\left(\pi_1(\phi_i),\pi_2(\phi_i)...\pi_n(\phi_i)\right)$ from the softmax layer, where $\pi_j(\phi_i)$ denotes the j-th output from the softmax probability distribution associated with the cluster $\phi_i$. Here we can equivalently express this as a conditional probability $\pi_j(\phi_i)=P(z=j|\theta=i)$. Then the weighted conditional probability is,
\begin{align}
    P(z=j|x,\boldsymbol{\mu},\boldsymbol{\sigma},\boldsymbol{\omega})=\sum_{i=1}^{k}{P(z=j|\theta=i)P(\theta=i|x,\boldsymbol{\mu},\boldsymbol{\sigma},\boldsymbol{\omega})}
\end{align}
Since each cluster is a Gaussian, we can substitute the conditional probability for the latent variable.
\begin{align}
    P(z=j|x,\boldsymbol{\mu},\boldsymbol{\sigma},\boldsymbol{\omega})&=\sum_{i=1}^{k}{P(z=j|\theta=i)\frac{\omega_iN(x|\mu_i,\sigma_i)}{\sum_{l=1}^{k}{\omega_lN(x|\mu_l,\sigma_l)}}} \\
    &=\sum_{i=1}^{k}{\pi_j(\phi_i)\frac{\omega_iN(x|\mu_i,\sigma_i)}{\sum_{l=1}^{k}{\omega_lN(x|\mu_l,\sigma_l)}}}
\end{align}
Then the predicted label $\tau$ of observation $x$ is essentially
\begin{align}
    \tau = \argmax_{j}\sum_{i=1}^{k}{\pi_j(\phi_i)\frac{\omega_iN(x|\mu_i,\sigma_i)}{\sum_{l=1}^{k}{\omega_lN(x|\mu_l,\sigma_l)}}} \quad \quad j=1,2,...n
\end{align}
where $\sum_{j=1}^{n}\pi_j(\phi_i)=1$.

\begin{figure}
    \vspace*{0.5cm}
    \centering
    \includegraphics[width=8cm]{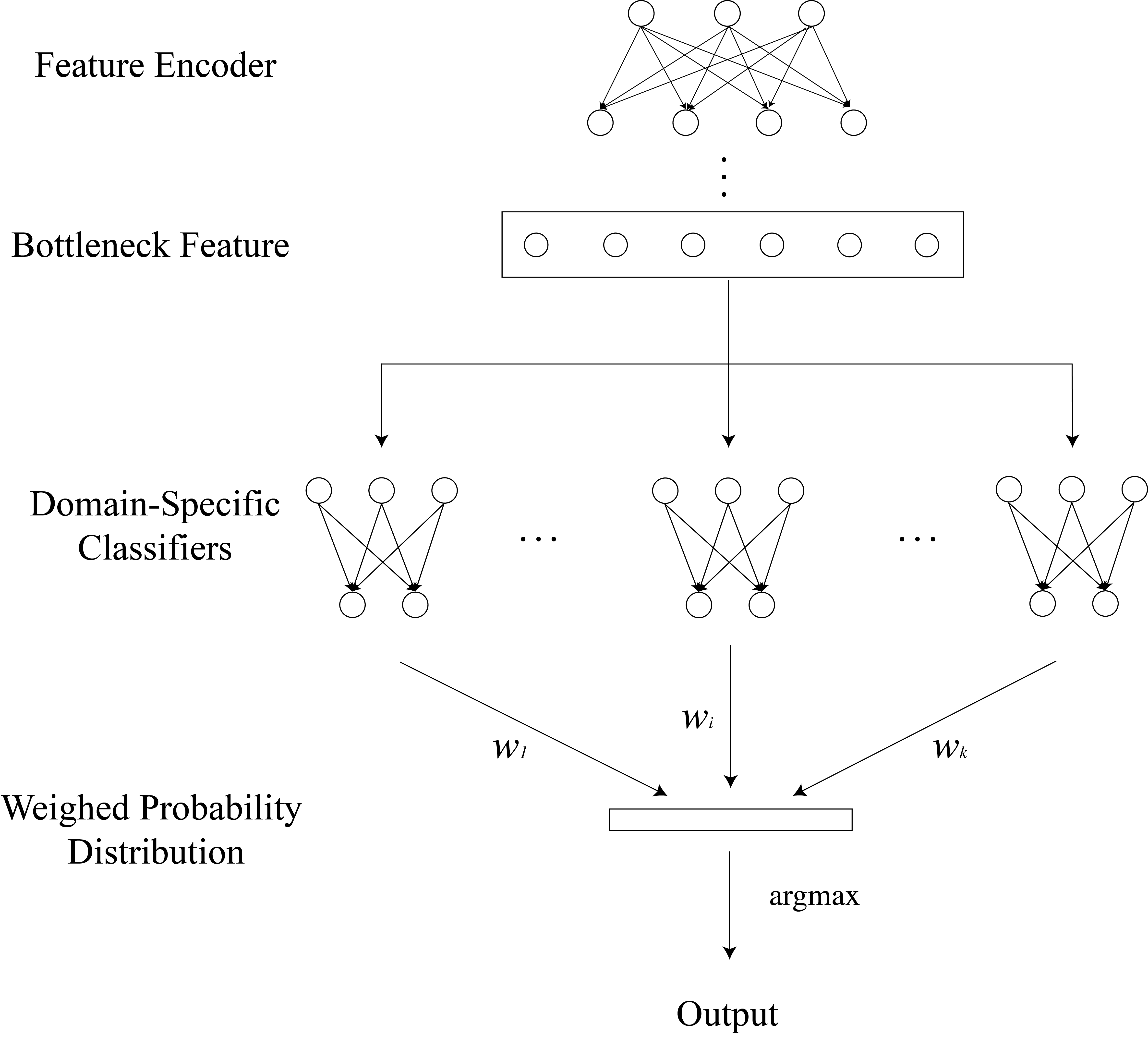}
    \caption{Architecture of Nonparametric Hierarchical Neural Network}
    \label{fig:my_label}
\end{figure}

Through its hierarchical structure, NHNN is able to extract common features across domains while preserving domain-specific characteristics. The structure can be adapted to a variety of neural networks such as CNN, RNN, and DNN. 

In the present study, we use a CNN-based NHNN for SER. The pipeline of the model is shown in Fig 2. The clustering of the audio is based on eGeMAPS feature vectors and the MFB features are passed into the feature encoder and emotion classifier. Both eGeMAPS and MFB features are commonly used in SER work and will be introduced in detail in Section 4.5. We use the feature encoder from the DCNN model to construct a shared bottleneck representation. 

In our preliminary clustering analysis, we find that, in each dataset, the clustering yields two major components that account for over 90\% of the utterances. Clusters with little data ($<10\%$ of the corpus) are therefore removed to ensure each cluster has sufficient data to fit the neural network. In experiments, the data are reassigned to existing cluster. 




\begin{figure}
    \centering
    \includegraphics[width=8cm]{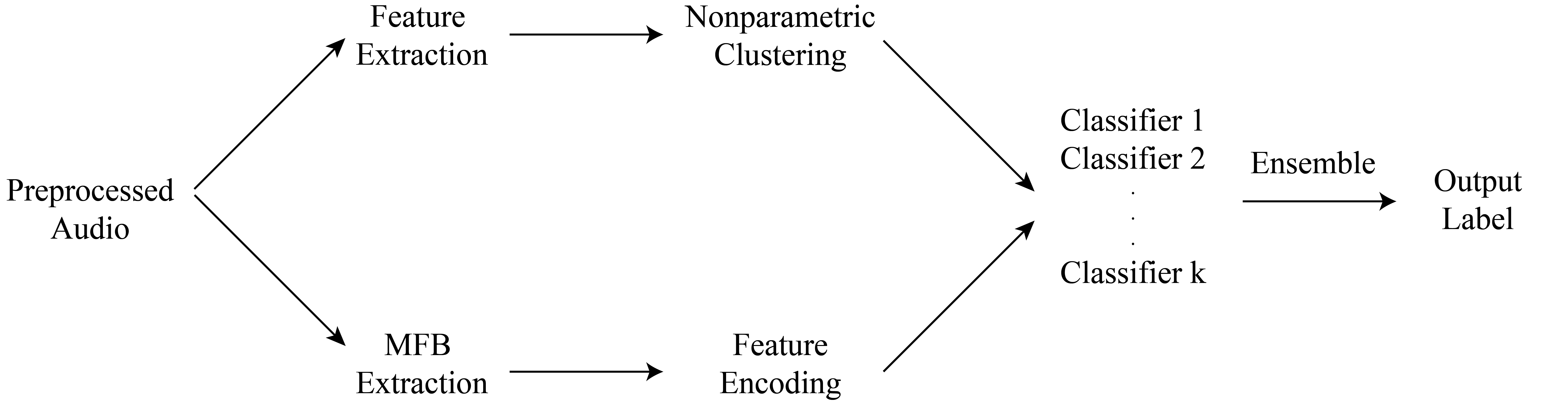}
    \caption{CNN-based NHNN implementation pipeline for speech emotion recognition}
    \label{fig:my_label}
\end{figure}

\subsection{Model Complexity}

The DCNN model has the least number of parameters (n\_parameters =175,619). The MTL-CNN model and the proposed NHNN FC model have a similar amount of parameters (n\_parameters =192,389 and 192,518 respectively). The proposed NHNN FC+Conv has more free parameters (n\_parameters=274,438) as the last layer of the encoder is tuned during training. The CNN-LSTM model has the most number of parameters (n\_parameters=530,691). In our experiments, we find the training time of each model is roughly proportional to the number of free parameters, where the CNN-LSTM takes the most time to train. 

\section{Datasets}

We select three datasets in the present study: IEMOCAP \cite{busso2008iemocap}, PRIORI Emotion \cite{khorram2018priori}, and PRIORI R21 . All three datasets are in English. IEMOCAP is a popular dataset produced in a lab, where both PRIORI Emotion and PRIORI R21 are in-the-wild datasets, which include phone calls with spontaneous speech and various background noises. We choose the three datasets so that the proposed model can be tested on both lab-recorded and in-the-wild datasets. Each dataset will be introduced in detail below.

\begin{table*}[h]
\vspace*{0.5cm}
\begin{center} {\footnotesize
\begin{tabular}{lccccccccc}
\hline Dataset
 & \multicolumn{1}{c}{Environment} & \multicolumn{1}{c}{Sampling} & \multicolumn{3}{c}{Speaker Level} & \multicolumn{4}{c}{Valence Rating}\\
 & &\multicolumn{1}{c}{Rate(kHz)} &
\multicolumn{1}{c}{All} & \multicolumn{1}{c}{Female} & \multicolumn{1}{c}{Male} & \multicolumn{1}{c}{All} & \multicolumn{1}{c}{Low} & \multicolumn{1}{c}{Medium}& \multicolumn{1}{c}{High}\\
\hline
{IEMOCAP} & Laboratory & 16 & 10 & 5 & 5 & 6,816 & 3,181 &1,641 & 1,994 \\[0ex]
{PRIORI Emotion} & Phone Calls & 8 & 19 & 12 &7&13,822 & 5,135& 6,579 & 2,108\\[0ex]
{PRIORI R21} & Phone Calls &  8 & 7 & 4& 3 & 2,677 & 1,228 &1,044 &405\\[0ex]

\hline
\end{tabular} }
\end{center}
\caption{Summary of datasets}
\label{turns}
\end{table*}

\subsection{IEMOCAP}

The Interactive Emotional Dyadic MOtion Capture Database (IEMOCAP) is a common lab-recorded multimodal dataset. Ten actors (five male and five female) performed a series of scripts or improvisational scenarios recorded over five sessions. The audios are recorded at a 48 kHz sampling rate and then downsampled to 16 kHz. The IEMOCAP dataset includes 10,039 audio segments (5,255 scripted utterances and 4,784 improvised utterances). The segments are annotated by two to four annotators on valence, activation, and dominance on a 5-point Likert Scale. 

\subsection{PRIORI Emotion}

The PRIORI Emotion dataset is an annotated subset of the larger PRIORI (Predicting Individual Outcomes for Rapid Intervention) bipolar mood dataset [cite]. It consists of phone calls from 19 subjects. The utterances are annotated on a 9-point Likert scale on valence and activation. The dataset includes 18,388 annotated audio segments.

\subsection{PRIORI R21}
PRIORI R21 is a multicultural subset of PRIORI. The PRIORI R21 dataset includes 9 subjects, 5 are native English speakers and 4 are native Arabic speakers. The countries of birth include Palestine, Lebanon, Saudi Arabia, Yemen, and the United States. Similar to PRIORI Emotion, the audio segments are annotated on a 9-point Likert scale on valence and activation. During annotation, non-English speech or empty segments are removed. The R21 dataset includes 4,148 annotated audio segments. In the present study, we exclude two subjects, one due to poor audio quality and the other due to a lack of audio segments (33 segments).

\subsection{Emotion Labels}

SER work often uses one of two emotion labels: categorical and dimensional labels. The categorical labels include emotions such as happy, sad, angry, etc. where dimensional labels typically include valence and activation. Valence refers to the perceived pleasantness of the speech and activation refers to the degree of arousal. Valence and activation are often scored on a Likert scale, which maps each utterance onto a 2-D valence-activation space.

Although much of the work on IEMOCAP uses categorical labels in classification tasks, in the present study, however, we choose to use dimensional valence scores because the PRIORI datasets are only annotated on dimensional scores. Previous research has also shown that dimensional labels are more consistently interpretable across datasets \cite{gideon2019improving, russell2003core}. 

We follow the same preprocessing procedure as Gideon et al. \cite{gideon2019improving} by converting each annotation of valence into three bins. The middle bin corresponds to medium valence with a rating of 3 for IEMOCAP and a rating of 5 for PRIORI Emotion and PRIORI R21. The other two bins correspond to valence ratings below and above the middle points.  We then create a final label for an utterance by aggregating over all evaluations for that utterance and identifying the most commonly selected (majority) bin.  We exclude all utterances without a majority bin. Our final datasets include 6,816 segments for IEMOCAP, 13,822 segments for PRIORI Emotion, and 2,677 segments for PRIORI R21. The breakdown of utterance annotations and demographics are shown in Table 1.

\subsection{Feature Extraction and Preprocessing}

All audio segments from the datasets are normalized to 0 dBFS using the Sox command line tool. We extract two sets of features for our model: eGeMAPS and MFB.

\textbf{eGeMAPS} – The eGeMAPS feature set is a commonly used feature set designed for SER. It includes 88 acoustic features such as energy, excitation, spectral, cepstral, etc. In the present study, the eGeMAPS feature vectors are extracted by the openSMILE toolkit with default parameters \cite{eyben2010opensmile,eyben2015geneva}.

\textbf{MFB} – We extract 40-dimensional MFB features with 25ms frame length and 10ms frame shift using the Librosa package \cite{mcfee2015librosa}. The MFB features are then z-normalized and padded to the same length during training to account for variations in audio lengths.

\section{Experimental Setup}

We designed two experiments to test the within-corpus and cross-corpus performance of NHNN against the baseline. In the present study, we train and test two variants of NHNN by freezing different layers of the DCNN model. The NHNN FC variant freezes all the convolutional layers and adapts the fully-connected layers to each cluster. The NHNN FC+Conv variant freezes one out of the two convolutional layers of the feature encoder.

In both experiments, the classifiers are trained on a single corpus. We randomly take out 1/4 of the training set to be the validation set. We implement both experiments with Pytorch. We use early stopping with a patience of 5 epochs and a maximum epoch of 50. The model with the best validation loss is selected. Training and testing is repeated with different random seeds. In both experiments, hyperparameters including batch size(32/64) and learning rate(0.001/0.0005/0.0001) are tuned for all models. We choose a batch size of 64 and a learning rate of 0.0001 for the final implementation.

For performance evaluation, we use the Unweighted Average Recall (UAR) as the performance metric. UAR ensures that each valence class is given equal weight at performance evaluation to account for possible data imbalance. A random prediction would result in a UAR of 0.333 in our experiments.

\subsection{Experiment 1}

In Experiment 1, we investigate the performance of NHNN on a single corpus. We train and test classifiers within-corpus for each of the three datasets. For each dataset, we use Leave-One-Subject-Out (LOSO) testing scheme and the performances are averaged over the number of subjects. We repeat the experiments with different random seeds and the final results are averaged. Since each model is tested with the same testing scheme, to evaluate the significance of performance difference, we use a paired t-test between results of the models models, where the UAR performance on each subject is paired from the LOSO testing results.

We also break down the UAR performances by groups including gender, language, and subject to compare the performances. We relate any performance discrepancies to differences between clusters to see if the clustering is improving the model performances on subsets/groups within the dataset.

\subsection{Experiment 2}

In Experiment 2, we investigate the cross-corpus performances by training classifiers on one dataset and testing their cross-corpus performances on each of the other two datasets. Since the PRIORI datasets and IEMOCAP are sampled with different rates, the IEMOCAP dataset is downsampled to 8 kHz in Experiment 2. Each model is trained 30 times and the results are averaged for comparison. Similar to experiment 1, we perform a paired t-test to compare the results.

\section{Results}

\subsection{Experiment 1}

The results of Experiment 1 are summarized in Table 2. The two proposed NHNN models achieve the best UAR performances in all three datasets when compared with other models. Between the two variants, the NHNN FC+Conv model shows superior UAR on all three datasets, compared to the NHNN FC variant. The Multi-task CNN model, when treating gender recognition as the auxiliary task, shows similar performance as the DCNN model. The CNN-LSTM model has a similar performance as the two NHNN models on IEMOCAP and PRIORI Emotion, where the differences are not statistically significant, but the CNN-LSTM performed poorly on PRIORI R21.

In the R21 dataset, the DCNN baseline model performance has a higher UAR for native English speakers (UAR=0.4017) than non-native speakers (UAR=0.3856) with statistical significance (p=0.0055). In NHNN FC, the performance gap decreases slightly from 1.61\% to 1.33\%, whereas in NHNN FC+Conv, it increases to 2.05\%. However, none of the changes in performance gaps is statistically significant.

The comparison analysis between the two gender groups shows that the performance gaps between male and female speech in PRIORI Emotion and PRIORI R21 shrink under our proposed models. In PRIORI Emotion, the UAR performance on female speech (UAR=0.4731) is 3.73\% higher than on male speech (UAR=0.4358). The gender-group UAR difference shrinks to 3.42\% under NHNN FC and 3.26\% under NHNN FC+Conv. In PRIORI R21, the UAR performance on male speech (UAR=0.3990) is 1.61\% higher than on female speech (UAR=0.3876). The gender-group UAR difference shrinks to 0.38\% under NHNN FC and -0.54\% under NHNN FC+Conv. No significant changes in gender-group UAR differences are observed for IEMOCAP when comparing two NHNN models and the DCNN model.

Table 3 shows the group-level performances improvement between female and male subjects against the DCNN model. In PRIORI Emotion and PRIORI R21, both variants of NHNN showed statistically significant performance boost on one of the gender groups but not the other, whereas the performance changes for the two gender groups is similar and both statistically significant on IEMOCAP. 

\subsection{Clustering Analysis}



The two mixture components have similar weights across three datasets, which roughly follow an 80/20 split. We first relate the clustering results to attributes of the audio segments. In particular, we look at the gender ratio, language ratio, subject audio distribution, and mood severity in each mixture. In the R21 dataset, we also analyze the proportion of audio segments from native English speakers in the mixtures. As gender and native-language are binary variables in our study, we simply compare the ratio of the two attributes in a dataset/mixture. The gender ratio is defined as the number of female audio segments divided by the number of male audio segments. The language ratio is calculated by dividing the number of audio segments from native English speakers from those from non-native speakers. As for subjects, there are great variations in the number of audio segments belonged to each subject in each dataset while the mixtures are of different sizes. To analyze the distributions of subject\_id, we divide the maximum number of audio segments belonged to a single subject by the minimum number as a surrogate of the dispersion of subject audio segments in a dataset/mixture. If the clustering is not based on a certain attribute, we would expect the relevant ratios to be similar in cluster 1, cluster 2, and the dataset as a whole.


In the R21 dataset, the clusters seem to be indicative of subjects' demographics, where gender ratio difference and native-language language difference are observed between the two clusters. In cluster 1, the Female/Male ratio is 0.65, and 1.53 in cluster 2. The English/Arabic as first language ratio is 2.24 in the first cluster and 0.95 in the second cluster. 

In the PRIORI Emotion dataset, the gender ratios are 1.96 and 0.69, respectively. Significant variations in subject\_id distributions are found in the two clusters. In the PRIORI Emotion dataset, the max/min subject audio segments ratio is 2.77 in cluster 1 and 9.00 in cluster 2, while the ratio is 1.56 in the V1 dataset as a whole.

As for IEMOCAP, the gender ratio follows an even split in each mixture and the dataset as a whole. We also did not observe any significant variations in the number of subject audio segments among clusters. It's likely that the clustering of IEMOCAP depends more on lower-level acoustic features than subject group differences.

The valence rating distributions are similar in all mixture components in each of the datasets.




\subsection{Experiment 2}

Table 4 shows the results of Experiment 2. The two NHNN models have better average performances than other models. In particular, NHNN FC+Conv outperforms DCNN and MTL-CNN on all cross-corpus tests, and CNN-LSTM when trained on IEMOCAP and PRIORI Emotion. The CNN-LSTM model performs relatively poorly on IEMOCAP and PRIORI Emotion against other models but has the best UAR performance when trained on PRIORI R21 and tested on the other two datasets. The differences are not statistically significant between CNN-LSTM and NHNN FC+Conv when trained on PRIORI R21 (p=0.1401 when tested on IEMOCAP and p=0.0986 when tested on PRIORI Emotion). Between the two variants, NHNN FC+Conv has on average better cross-corpus performances than NHNN FC model.

\begin{table*}[h]
\begin{center} 
\setlength{\tabcolsep}{12pt}
\newcolumntype{P}[1]{>{\centering\arraybackslash}p{#1}}
\renewcommand{\arraystretch}{1.2}
\begin{tabular}{cP{1.5cm}P{1.5cm}P{1.5cm}:P{1.5cm}P{1.5cm}}
\hline 
\multirow{3}{*}{Dataset}&\multicolumn{4}{c}{Performance (UAR)}\\\cline{2-6}
&\multicolumn{1}{P{1.5cm}}{\centering  DCNN}&\multicolumn{1}{P{1.5cm}}{\centering MTL \\ CNN}&

\multicolumn{1}{P{1.5cm}}{\centering CNN-LSTM } &
\multicolumn{1}{:P{1.5cm}}{\centering NHNN \\FC} & \multicolumn{1}{P{1.5cm}}{\centering NHNN \\FC+Conv}\\ \hline

IEMOCAP  &0.5452  & 0.5482 &0.5517&$0.5551^{*}$& $\textbf{0.5568}^{*}$  \\
PRIORI Emotion &0.4594  & 0.4450 &0.4639& $0.4654^{*}$& $\textbf{0.4685}^{*}$ \\
PRIORI R21  & 0.3925 &0.3993&  0.3876 & 0.4001 & $\textbf{0.4071}^{\dag}$ \\

\hline
\end{tabular} 
\end{center}
\caption{Within-corpus test performances of models. Best UAR performances are bolded. * indicates the performance is better than DCNN and MTL-CNN with p<0.05. \dag indicates the performance is better than CNN-LSTM with p<0.05.}
\label{turns}
\end{table*}

\begin{table*}[h]
\begin{center}
\setlength{\tabcolsep}{12pt}
\newcolumntype{P}[1]{>{\centering\arraybackslash}p{#1}}
\renewcommand{\arraystretch}{1.2}
\begin{tabular}{cP{2cm}P{2cm}P{2cm}}
\hline 
\multicolumn{1}{c}{Dataset}
&\multicolumn{1}{P{2cm}}{Model}&
\multicolumn{1}{P{2cm}}{\centering $\Delta$UAR$_{male}$} & \multicolumn{1}{P{2cm}}{\centering $\Delta$UAR$_{female}$}\\ \hline

\multirow{ 2}{*}{IEMOCAP} & FC  &  \textbf{0.0096} & \textbf{0.0101} \\
& FC+Conv & \textbf{0.0095} & \textbf{0.0136}\\
\multirow{ 2}{*}{PRIORI Emotion} &  FC & \textbf{0.0121}& 0.0074     \\
& FC+Conv & \textbf{0.0080}  & 0.0048 \\
\multirow{ 2}{*}{PRIORI R21}  & FC & 0.0033& \textbf{0.0109} \\
& FC+Conv  & 0.0050 & \textbf{0.0218} \\

\hline
\end{tabular} 
\end{center}
\caption{Gender group-level average UAR performance change compared with DCNN model. Statistically significant improvements (p<0.05) are bolded}

\label{turns}
\end{table*}

\begin{table*}[h]
\begin{center} 
\setlength{\tabcolsep}{12pt}
\newcolumntype{P}[1]{>{\centering\arraybackslash}p{#1}}
\renewcommand{\arraystretch}{1.2}
\begin{tabular}{cP{2.5cm}P{2cm}P{2cm}P{2cm}}
\hline 
\multicolumn{1}{c}{Dataset}
&\multicolumn{1}{P{2.5cm}}{Ratio}&
\multicolumn{1}{p{2cm}}{\centering Cluster 1} & \multicolumn{1}{p{2cm}}{\centering Cluster 2} & \multicolumn{1}{p{2cm}}{\centering Total}
\\ \hline

\multirow{ 2}{*}{IEMOCAP} & Gender  & 1.05 & 1.01 &1.04  \\
& Subject  & 1.60 & 2.13 & 1.35 \\
\hline
\multirow{ 2}{*}{PRIORI Emotion} &  Gender & 1.96 & 0.69 & 1.67   \\
& Subject & 2.77 & 9.00 & 1.56 \\
\hline
\multirow{ 3}{*}{PRIORI R21}  & Gender & 0.65 & 1.53 & 1.28 \\
&language & 2.44 & 0.95 & 1.15 \\
& Subject  & 13.22 & 85.25 & 17.13 \\

\hline
\end{tabular} 
\end{center}
\caption{Key ratios among datasets and clusters}
\label{turns}
\end{table*}


\begin{table*}[h]
\begin{center} 
\setlength{\tabcolsep}{12pt}
\newcolumntype{P}[1]{>{\centering\arraybackslash}p{#1}}
\renewcommand{\arraystretch}{1.2}
\begin{tabular}{ccP{1cm}P{1cm}P{1cm}:P{1.2cm}P{1.2cm}}
\hline 
\multirow{3}{*}{Train On}&\multirow{3}{*}{Test On}&\multicolumn{5}{c}{Performance (UAR)}\\\cline{3-7}
&&\multicolumn{1}{P{1cm}}{\centering DCNN}&
\multicolumn{1}{P{1cm}}{\centering MTL \\ CNN}&
\multicolumn{1}{P{1cm}}{\centering CNN-LSTM}&
\multicolumn{1}{:P{1.2cm}}{\centering NHNN \\FC} & \multicolumn{1}{P{1.2cm}}{\centering NHNN \\FC+Conv}\\ \hline

IEMOCAP & PRIORI Emotion & 0.3572 &0.3742 & 0.3622 & $0.3831^{*}$ & $\textbf{0.3873}^{*\dag}$  \\
IEMOCAP & PRIORI R21 & 0.4002 & 0.3836 &  0.4027 & 0.3989 & $\textbf{0.4059}^{*\dag}$ \\
PRIORI Emotion & IEMOCAP & 0.4056 & 0.4003 & 0.3873 & $\textbf{0.4084}^{*\dag}$ & $0.4083^{\dag}$ \\
PRIORI Emotion & PRIORI R21 & 0.4488 & 0.4353 & 0.4399& $\textbf{0.4562}^{*\dag}$& $0.4540^{*\dag}$ \\
PRIORI R21 & IEMOCAP & 0.3441 &0.3471 & \textbf{0.3572} & $0.3444^{*}$& $0.3508^{*}$\\
PRIORI R21 & PRIORI Emotion & 0.3499&0.3611&\textbf{0.3752} & $0.3560^{*}$&$0.3684^{*}$\\

\hline
\end{tabular} 
\end{center}
\caption{Cross-corpus performances of models. Best UAR performances are bolded. * indicates the performance is better than DCNN and MTL-CNN with p<0.05. \dag indicates the performance is better than CNN-LSTM with p<0.05.}
\label{turns}
\end{table*}

\section{Discussion}

In the present study, we introduce the Nonparametric Hierarchical Neural Network, which learns latent structures from the dataset with a Dirichlet Process Mixture Model to build domain-specific classifiers from a shared feature encoder. 

The two experiments we conducted show that NHNN, despite being a lightweight model, has superior within-corpus performance than DCNN and MTL-CNN. Its cross-corpus test performance is on average better than DCNN, MTL-CNN and CNN-LSTM, which indicates a better generalizability on unseen data. The performance increase over the DCNN, which uses the same feature encoder as the NHNN model, is observed for both lab-produced and in-the-wild datasets. In addition, NHNN FC+Conv shows a slightly better average performance than NHNN FC in both experiments. This is likely because NHNN FC+Conv allows for a more versatile classifier with more free parameters. 

Comparing results across two experiments, we notice that the model trained on PRIORI Emotion and tested on PRIORI R21 has a significantly better performance than within-corpus UAR of PRIORI R21. This suggests that the low R21 within-corpus performance relative to PRIORI Emotion is likely due to a lack of training samples. 

In the cluster analysis, we show that the clustering criteria vary depending on the dataset, which may be a combination of higher-level attributes or lower-level acoustic features. Our subsequent analysis also confirms that the performance improvement of the NHNN models tend to be different depending on domains. In Experiment 1, we found that the performance improvement on one gender is significantly better than the other when the dataset has gender imbalances in utterances and the clustering is partially based on gender. When the clustering is not based on gender, and the gender ratio is close to 1, as is the case for IEMOCAP, the model showed a similar performance increase for both genders. This implies that the clustering-based approach has an effect on the model performance on some selective gender-related data with local characteristics that are not properly accounted for in a general CNN model when there is a gender imbalance. Through clustering, we build customized classifiers for each mixture component to account for these variations, while using a shared feature encoder to preserve group-invariant characteristics. Compared with a typical blackbox machine learning algorithm, our proposed model and subsequent analysis provide a more intuitive and explainable approach.

In addition to performance, compared with multi-tasking learning, which has a similar training time and model complexity, our model is more flexible as it does not require any group labels for training and achieves better within-corpus and cross-corpus performances with statistical significance.

Lastly, as much of the emotional experiences, expressions and perceptions are biologically, socially and culturally dependent, the ability to understand and account for domain-specific characteristics is crucial in building equitable SER systems for in-the-wild speech by speakers from all backgrounds. The NHNN model has the potential of accounting for variations and reducing model biases.

Future studies can explore in detail how the clustering of acoustic features is related to sources of variation. This will help us better understand domain-specific features in speech emotion expressions. In our experiments, we used a standard DPMM with a full set of eGeMAPS features. Future studies can conduct feature selection and regularization techniques to test different variants of NHNN.

\section{Conclusion}

In this study, we propose the Nonparametric Hierarchical Neural Network to account for variations in emotional expressions in speech. In our experiments, we show the proposed model surpass state-of-the-art models on most of within-corpus and cross-corpus SER tests.

\bibliography{BibliographyFile}
\bibliographystyle{plain}
\end{document}